\documentclass{article}
\usepackage{spconf,amsmath,graphicx,amssymb}
\usepackage{caption}
\usepackage{amsfonts}
\usepackage{subcaption}
\usepackage{fancyhdr}
\usepackage{hyperref}

\title{Image Completion and Extrapolation with Contextual Cycle Consistency}
\name{Sai Hemanth Kasaraneni, Abhishek Mishra}
\address{Samsung Research Institute - Noida, India\\
k.saihemant@samsung.com, m.abhishek@samsung.com}
\begin{document}
%
\maketitle
\begin{abstract}
Image Completion refers to the task of filling in the missing regions of an image and Image Extrapolation refers to the task of extending an image at its boundaries while keeping it coherent. Many recent works based on GAN have shown progress in addressing these problem statements but lack adaptability for these two cases, i.e. the neural network trained for the completion of interior masked images does not generalize well for extrapolating over the boundaries and vice-versa. In this paper, we present a technique to train both completion and extrapolation networks concurrently while benefiting each other. We demonstrate our method's efficiency in completing large missing regions and we show the comparisons with the contemporary state of the art baseline.
\end{abstract}
\section{Introduction}
\label{sec:intro}
Image Completion and Extrapolation (often referred to as image inpainting and outpainting respectively) becomes extremely difficult when the area to be filled is high. It is often required to complete high structural objects and reproduce the textural characteristics of the image. Previously, many patch-based image synthesis methods \cite{Barnes:2009:PAR,Darabi12:ImageMelding12,microsoftcompletion} were proposed to complete missing regions in images. But those methods lack the capability to generate new content from the available context of the image. They also fail to complete objects with high-level structural characteristics.

Recently, Deep Learning based methods \cite{Pathak_2016_CVPR},\cite{yeh2017semantic},\cite{IizukaSIGGRAPH2017},\cite{Yu_2018_CVPR},\cite{portrait} made significant progress on filling missing regions of images based on context. All these methods are based on Generative Adversarial Networks (GAN) \cite{goodfellow2014generative} which employ adversarial training to train generator and discriminator. \cite{Pathak_2016_CVPR} trained a convolution neural network (CNN) based encoder-decoder to encode the context information of masked image and utilize it to generate content over the masked regions. \cite{yeh2017semantic} demonstrated an iterative approach to find the appropriate noise prior of GAN trained on real image dataset for finding semantically closest possible complete image to the masked input image. But it comes at a cost of significant inference time due to the iterative search. \cite{IizukaSIGGRAPH2017,Yu_2018_CVPR,portrait} employed two discriminators trained along with the completion network, a global discriminator and a local discriminator. Global discriminator was trained to distinguish inpainted images from real images while local discriminator was trained to distinguish the network completed regions from the random patches of real images. However, these methods incur heavy distortions and blurriness when the area of the regions to be filled is high.

Despite the fact that the image extrapolation is a subcase of image completion where the masked regions exist at the boundaries of the image, it is hard to extrapolate at all the boundaries without being explicitly trained because of the long distance textural dependencies.

\section{Preliminaries}
\label{sec:pre}

GANs \cite{goodfellow2014generative} are the deep generative models that contain two networks that compete with each other, Generator G and Discriminator D and are trained adversarially. The Generator is optimized to generate images that are distributed similar to the real image dataset distribution so that the discriminator will not be able to distinguish them. The discriminator is optimized concurrently to distinguish between real and synthetic images efficiently. Their objectives are contrary and both the optimizations occur concurrently over a loss function like playing a minimax game.

\begin{equation}
  \begin{split}
  \min_{G} \max_{D} \mathcal{L}(G,D) = \mathbb{E}_{x \sim p_data(x)}[\log D(x)] +\\
   \mathbb{E}_{z \sim p_z(z)}[\log (1-D(G(z)))]\label{eq1}
  \end{split}
\end{equation}

\begin{figure*}[htbp]
\centerline{\includegraphics[width = \linewidth]{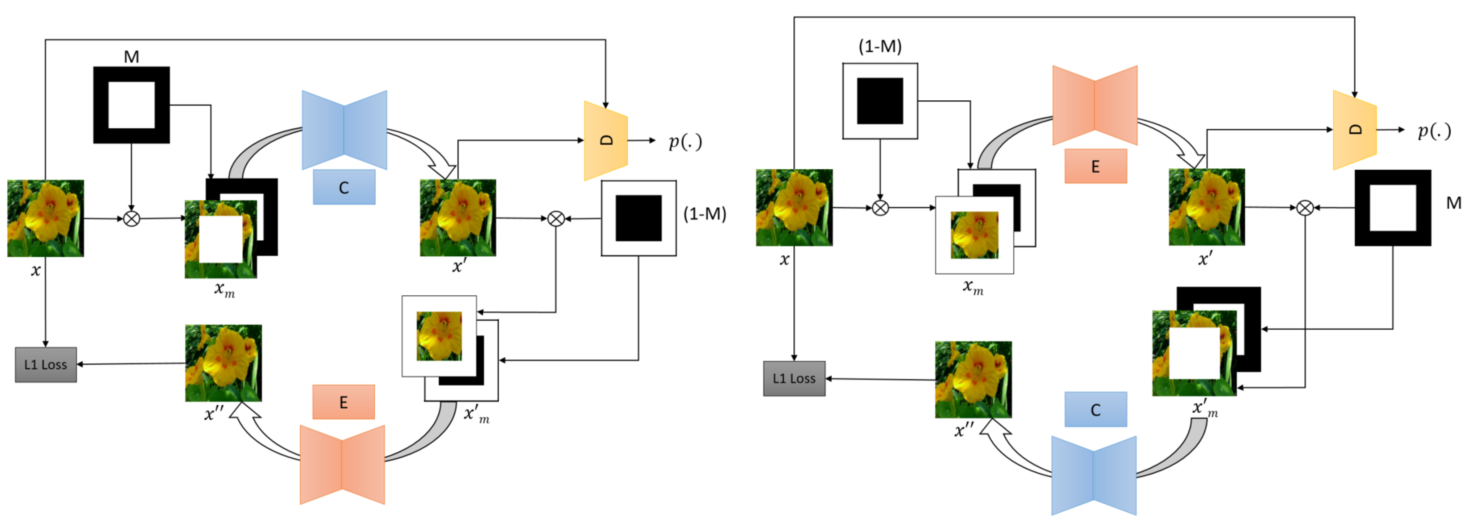}}
\caption{\scriptsize The left figure shows the architecture of the forward cycle and the right figure shows the architecture of the backward cycle. Forward cycle performs completion and then extrapolation and vice-versa. Both the cycle losses are optimized together.}
\label{fig1}
\end{figure*}

Eq.\eqref{eq1} is the loss function for a typical GAN. Cyclegan \cite{Zhu_2017_ICCV} used two such generators and discriminators to perform unpaired image to image translation where one generator translates a domain 'A' image to domain 'B' while other generator translates a domain 'B' image back to domain 'A'. Problem domains like image to image translation, inpainting and extrapolation often require to reproduce the existing semantics and complex textures from subject image.

\cite{Zhu_2017_ICCV} employed cycle consistency loss to achieve transitivity in unpaired image to image domain translation. They achieved semantically plausible results and were able to preserve the context of the image throughout the cycle. Motivated by the encouraging results of \cite{Zhu_2017_ICCV}, we use a similar approach to train completion and extrapolation networks together to maintain contextual cycle consistency.

\section{Method}
\label{sec:meth}

Our architecture contains a completion network 'C', an extrapolating network 'E' and a discriminator 'D'. The completion network is trained to fill the missing portions of the input image and the extrapolating network is trained simultaneously to extrapolate the input image while the discriminator is trained to differentiate the distributions of synthetic images and real images.

\textbf{Similarity with existing baselines:}\cite{IizukaSIGGRAPH2017},\cite{Yu_2018_CVPR},\cite{portrait} used a global discriminator and a local discriminator along with the completion network to complete masked images. Global discriminator was a normal discriminator that discriminates inpainted images from the real images. Local discriminator was trained to discriminate between random patches of real images and only the inpainted portions of the completed images. We use only one global discriminator which distinguishes whole inpainted images from the real images. Instead of training another local discriminator, we train the extrapolation network to extrapolate based on the inpainted region and to reconstruct back the original image. We define this procedure as forward cycle. Both the networks are optimized over the reconstruction error, driving them to propagate the textural characteristics as well as the context of the image throughout the cycle. We aim to propagate the context of the image between inside masked and outside masked images via the completion and extrapolation networks by driving context-based content creation. When both the networks are successful in propagating the context of the image through generated regions, the final reconstructed image or the cycle output will be semantically and visually similar to the original image or contextually consistent with the original image. We also perform training on the reverse cycle and optimize both networks over the loss functions of both the cycles. In section \ref{sec:exp}, we demonstrate that our approach is suitable for semantically filling large missing regions in the image. Fig.~\ref{fig1} illustrates the training procedure of our architecture.

We define a binary mask matrix "M" that has the same spatial resolution as the input image and is filled with ones within surrounding boundaries filled with zeros. We experimented with masks of square shapes of area chosen randomly in the range 25 to 35 percent of the image. This is a critical parameter as both the networks require sufficient unmasked content to extract the context of the image. The pixels with values '1' denote the region that is to be painted synthetically and '0's denote the unmasked regions in the image. Let the input image be 'x' which is drawn from the training dataset. The inside masked image will be $(\textbf{1}-M) \odot x$, where $\textbf{1}$ stands for the unity matrix of same size as that of mask.

The masked image, concatenated along with the binary mask representing the region to be filled is given as input to the completion network to get the inpainted image. We take the adversarial loss over the inpainted region to converge the completed image to be perceptually plausible. This perceptual loss ensures that the inpainted image lies near the real data manifold. The perceptual loss function associated with the inpainted image will be:

\begin{equation}
  \mathcal{L}_{\text{adv}}(C,D,M) = \mathbb{E}[\log D(x) + \log (1-D(C((1-M)\odot x))]\label{eq2}
\end{equation}

We also consider the contextual loss or the fidelity loss incurred in the completion of the image to preserve the fidelity in completion. It is computed as the L1 norm of the difference of the pixel values of the original image and the output image over the completed region :

\begin{equation}
  \mathcal{L}_{\text{ctx}}(C,M) = \lVert M\odot(C((1-M)\odot x) - x) \rVert\label{eq3}
\end{equation}

The inpainted image will be given as input to extrapolation network after inverse masking and concatenating with the inverse of the defined mask 'M'. We compute L1 reconstruction loss between the final output and the original input image and optimize both the networks over this loss to preserve the context of the original image throughout the cycle.

\begin{equation}
  \mathcal{L}_{\text{rec}}(C,E,M) = \lVert (1-M)\odot(E(M\odot C((1-M)\odot x)) - x) \rVert\label{eq4}
\end{equation}

In the above equations $\lVert . \rVert$ denotes L1 - norm and $\odot$ denotes the element wise multiplication of the matrices. We optimize over the adversarial loss of the extrapolator network as well. The total forward cycle loss $\mathcal{L}_{\text{cyc}}(C,E,M)$ would be the weighted sum of the above three losses.

\begin{equation}
\begin{split}
  \mathcal{L}_{\text{cyc}}(C,E,M) = \mathcal{L}_{\text{adv}}(C,D,M) + \alpha.\mathcal{L}_{\text{ctx}}(C,M) +\\ \beta.\mathcal{L}_{\text{rec}}(C,E,M)\label{eq5}
\end{split}
\end{equation}

Similarly we compute losses for the reverse cycle as well i.e. we give outside masked input to extrapolation network and further to the completion network after inverse masking and finally compute the total loss for the reverse cycle $\mathcal{L}_{\text{cyc}}(C,E,\textit{\textbf{1}-}M)$.

\begin{equation}
\begin{split}
  \mathcal{L}_{\text{cyc}}(C,E,\textit{\textbf{1}}-M) = \mathcal{L}_{\text{adv}}(E,D,\textit{\textbf{1}}-M) +\\ \alpha.\mathcal{L}_{\text{ctx}}(E,\textit{\textbf{1}}-M) +\\ \beta.\mathcal{L}_{\text{rec}}(C,E,\textit{\textbf{1}}-M)\label{eq6}
\end{split}
\end{equation}

Now the completion, extrapolation and discriminator networks are optimized over the combined loss function.

\begin{equation}
  \begin{split}
  \min_{C,E} \max_{D}L(C,E,D,M) = \min_{C,E} \max_{D}[\mathcal{L}_{\text{cyc}}(C,E,M) +\\ \mathcal{L}_{\text{cyc}}(C,E,\textit{\textbf{1}}-M)\label{eq7}]
\end{split}
\end{equation}

\begin{figure}[h]
\centerline{\includegraphics[scale=0.5]{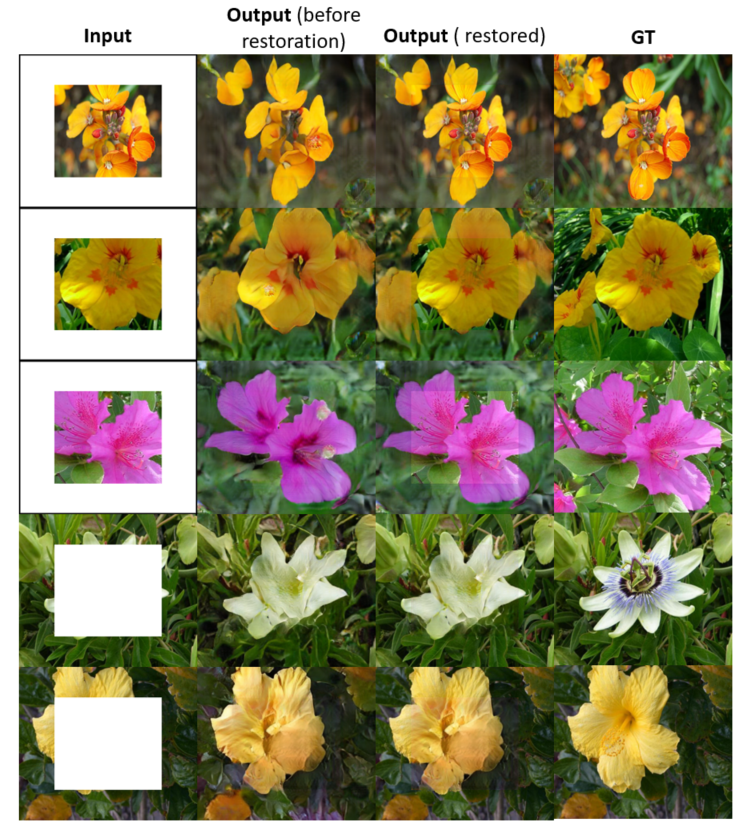}}
\caption{\scriptsize Completion and Extrapolation results on the Flowers102 test dataset. First three rows demonstrate extrapolation and the last two rows demonstrate the completion. First column images are the masked inputs. Second column images are the outputs from extrapolation and completion networks respectively before restoring the unmasked data from the input image. Third column are the outputs after restoring the unmasked data. Fourth column are the ground truth images \textbf{GT}.}
\label{fig2}
\end{figure}

\section{Experiments}
\label{sec:exp}

The completion and extrapolation network architectures are derived from the generator architecture in cyclegan. We concur with \cite{IizukaSIGGRAPH2017} in using dilated convolutions in the middle layers of the generator networks to accommodate exponential growth of the receptive field.

\textbf{Mask M: } Our architecture is suitable for mask shapes of any enclosed figure. Nevertheless, for all of the comparisons, we trained and tested our models with an enclosed mask of square shape. The area of the inside masked region carries a crucial role in the convergence of the model. A large amount of masked region imposes large area to be filled by the completion network and decelerates its convergence. On the contrary, small masked regions result in large inverse masked inputs to extrapolation network and would require it to reproduce the structural and textural characteristics based on the limited available content in the image. In our experiments, we set the mask area randomly in the range of 25 to 35 percent of the input image area and observed the stable convergence in both the networks.

\begin{figure}[h]
\centerline{\includegraphics[scale=0.45]{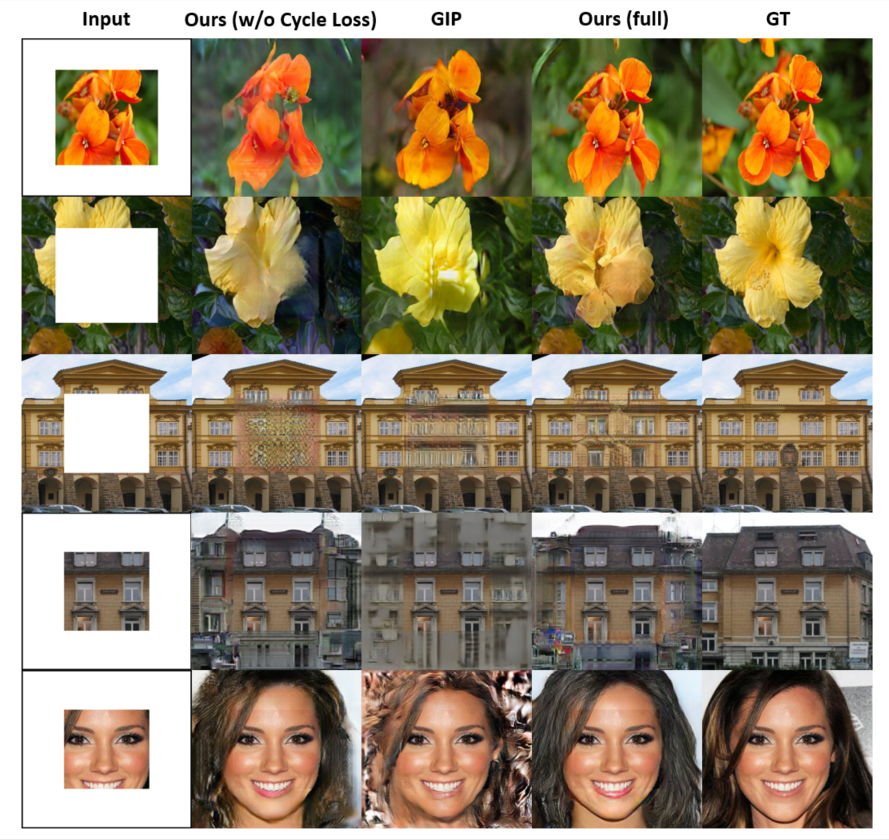}}
\caption{\scriptsize Visual comparison of completion and extrapolation results. First column are the masked inputs. Second column are outputs of completion and extrapolation networks trained independently without cycle reconstruction loss on inside masked and outside masked images respectively. Third column are the inference outputs of GIP \cite{Yu_2018_CVPR}. We inferred their pre-trained model on CelebA-HQ dataset. For results on flowers and facades, we trained their model using their official implementation. Fourth column are the outputs from our completion and extrapolation networks. Fifth column images are Ground Truth}
\label{fig3}
\end{figure}

\begin{figure}[h]
\centerline{\includegraphics[scale=0.5]{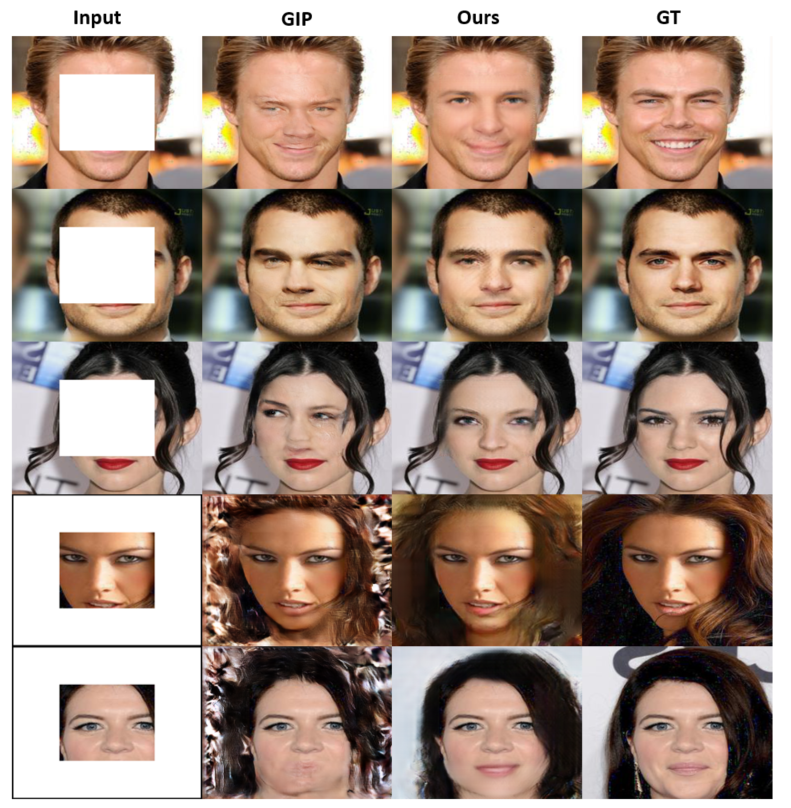}}
\caption{\scriptsize Visual Comparison of Face Completion and Extrapolation results on the CelebA-HQ dataset. First column images are the masked inputs. Second column are the outputs from \textbf{GIP} \cite{Yu_2018_CVPR}. We used their official repository and pre-trained model on the CelebA-HQ dataset for inference. Third column are our results using completion and extrapolation networks. Fourth column are the ground truth images \textbf{GT}. [Notice that \textbf{GIP} failed to reconstruct hair at the boundaries. Best viewed when zoomed-in]}
\label{fig4}
\end{figure}

\textbf{$\alpha$ and $\beta$ : } $\alpha$ and $\beta$ are the hyperparameters to impose relative importance between and perceptual and contextual losses formulated in Eq.\ref{eq5} and Eq.\ref{eq6} . As proposed in \cite{Zhu_2017_ICCV}, we set $\alpha$ = $\beta$ = 10 in all our experiments.

As suggested in \cite{IizukaSIGGRAPH2017}, general inpainting and outpainting methods restore the pixel values at the unmasked region from the input image to the output so as to avoid changes in the existing data. We manifest that our model can replicate the existing data in the input while completing and extrapolating images. Fig.\ref{fig2} shows the inpainted and outpainted images by the completion and extrapolation networks trained on Oxford flowers 102 \cite{nilsback2008automated} dataset. We presented the immediate outputs from the completion and extrapolation networks and compared them with the restored outputs. It is evident from this figure that our model is able to reproduce the existing shapes and textural characteristics of the input image.

\textbf{Qualitative Analysis :} We evaluate our model's performance on CelebA-HQ faces \cite{karras2017progressive}, Oxford 102 Category Flowers \cite{nilsback2008automated} and CMP Facade \cite{Tylecek13} datasets. In all our experiments we trained with input images of resolution 256 $\times$ 256 and no post-processing or image blending techniques have been applied after image completion or extrapolation. We emphasize that the cycle consistency loss (Eq.\ref{eq4}) is critical in modeling the transitivity and convergence of the model. Fig.\ref{fig3} shows the visual comparison of completion and extrapolation of masked images with and without using cycle consistency loss. We observed that the completion network was not able to converge on celebrity faces dataset without cycle consistency loss. We observed similar characteristics for the extrapolation of facades. Notice that the GIP \cite{Yu_2018_CVPR} particularly fails to give comprehensible results for outpainting of images at boundaries. Fig.\ref{fig4} shows more inferences on celebrity faces inpainting and outpainting. We used the official repository and pre-trained model on CelebA-HQ dataset of GIP \cite{Yu_2018_CVPR}.

\begin{table}[!t]
 \scriptsize
\caption{\scriptsize Average PSNR (in dB) values on test sets of different datasets. We considered both outpainting and inpainting results for each test image. Higher value means better reconstruction.\label{table_psnr}}
\centering
\begin{tabular}{cccc}\\\hline
   & \textbf{Ours (w/o cycle loss)} & \textbf{GIP \cite{Yu_2018_CVPR}} & \textbf{Ours (full)} \\
  CelebA-HQ & 22 & 22.3 & \textbf{28.6} \\
  Flowers102 & 23.1 & \textbf{25.8} & 25.2\\
  Facades & 16 & 15.8 & \textbf{19.3}

 \\\hline

\end{tabular}
\end{table}

\textbf{Quantitative Analysis :} There are several metrics like PSNR, Inception Score \cite{NIPS2016_6125} to evaluate image inpainting and outpainting techniques but are not flawless. However, we report PSNR values on test sets of the mentioned datasets and compare with that of \cite{Yu_2018_CVPR} in Table.[1]

\section{Conclusion}
In this paper, we first discussed the problem of compatibility between image inpainting and outpainting models. We then proposed a technique to train both completion and extrapolation networks while benefitting each other. Several comparisons with qualitative and quantitative analysis were made with the contemporary state of the art solution and our approach outperformed them. Future work will be focused on further improving the consistency loss to make it more appropriate to image completion.

\bibliographystyle{IEEEbib}
\bibliography{refs}

\end{document}